\documentclass[sigconf]{acmart}

\renewcommand\footnotetextcopyrightpermission[1]{} 
\usepackage{booktabs} 
\usepackage[usestackEOL]{stackengine}
\strutlongstacks{T}

\usepackage{tikz}
\usetikzlibrary{positioning,shapes,shadows,arrows}
\usepackage{xcolor}

\setcopyright{none}

\begin{document}
\title[Neural Network Architecture Search with dCGP for Regression]{Neural Network Architecture Search with Differentiable Cartesian Genetic Programming for Regression}

\author{Marcus M\"{a}rtens}
\orcid{0000-0003-1950-7111}
\affiliation{%
  \institution{European Space Agency}
  \city{Noordwijk} 
  \country{The Netherlands} 
}
\email{marcus.maertens@esa.int}

\author{Dario Izzo}
\affiliation{%
  \institution{European Space Agency}
  \city{Noordwijk} 
  \country{The Netherlands} 
}
\email{dario.izzo@esa.int}


\begin{abstract}
The ability to design complex neural network architectures which enable effective training by stochastic gradient descent has been the key for many achievements in the field of deep learning.
However, developing such architectures remains a challenging and resource-intensive process full of trial-and-error iterations. All in all, the relation between the network topology and its ability to model the data remains poorly understood.
We propose to encode neural networks with a differentiable variant of Cartesian Genetic Programming (dCGPANN) and present a memetic algorithm for architecture design: local searches with gradient descent learn the network parameters while evolutionary operators act on the dCGPANN genes shaping the network architecture towards faster learning.
Studying a particular instance of such a learning scheme, we are able to improve the starting feed forward topology by learning how to rewire and prune links, adapt activation functions and introduce skip connections for chosen regression tasks.
The evolved network architectures require less space for network parameters and reach, given the same amount of time, a significantly lower error on average.
\end{abstract}

%
%
%
%


\maketitle

\section{Introduction}
\label{sec:intro}

The ambition of artificial intelligence (AI) is to develop artificial systems that exhibit a level of intelligent behaviour competitive with humans. It is thus natural that many research in AI has taken inspiration from the human brain~\cite{hassabis2017neuroscience}. The general brain was shaped by natural evolution to give its owner the ability to learn: new skills and knowledge are acquired during lifetime due to exposure to different environments and situations. This lifelong learning is in stark contrast to the machine learning approach, where typically only weight parameters of a static network architecture are tuned during a training phase and then left frozen to perform a particular task.

While exact mechanisms in the human brain are poorly understood, there is evidence that a process called \emph{neuroplasticity}~\cite{draganski2004neuroplasticity} plays an important role, which is described as the ability of neurons to change their behaviour (function, connectivity patterns, etc.) due to exposure to the environment~\cite{rauschecker1981effects}. These changes manifest themselves as alterations of the physical and chemical structures of the nervous system.

Inspired by the idea of the neuroplastic brain, we propose a differentiable version of Cartesian Genetic Programming (CGP)~\cite{miller2011cartesian} as a direct encoding of artificial neural networks (ANN), which we call dCGPANN. Due to an efficient automated backward differentiation, the loss gradient of a dCGPANN can be obtained during fitness evaluation with only a negligible computational overhead. Instead of ignoring the gradient information, we propose a memetic algorithm that adapts the weights and biases of the dCGPANN by backpropagation. The performance in learning is then used as a selective force for evolution to incrementally improve the network architecture. We trigger these improvements by mutations on the neural connections (rewirings) and the activation functions of individual neurons, which allows us to navigate the vast design space of neural network architectures up to a predetermined maximum size.

To evaluate the performance of our approach, we evolve network architectures for a series of small-scale regression problems. Given the same canonical feed forward neural network as a starting point for each individual challenge, we show how complex architectures for improved learning can be evolved without human intervention. 

The remainder of this work is organized as follows: Section~\ref{sec:related} relates our contribution to other work in the field of architecture search and CGP applied to artificial neural networks. Section~\ref{sec:dcgp} gives some background on CGP, introduces the dCGPANN encoding and explains how its weights can be trained efficiently. In Section~\ref{sec:experiments} we outline our experiments and describe the evolutionary algorithm together with our test problems. Results are presented in Section~\ref{sec:results} and we conclude with a discussion on the benefits of the evolved architectures in Section~\ref{sec:discussion}.

\section{Related Work}
\label{sec:related}

This section briefly explains how this work is related to ongoing research like genetic programming, neural network architecture search, neuro-evolution, meta-learning and similar.

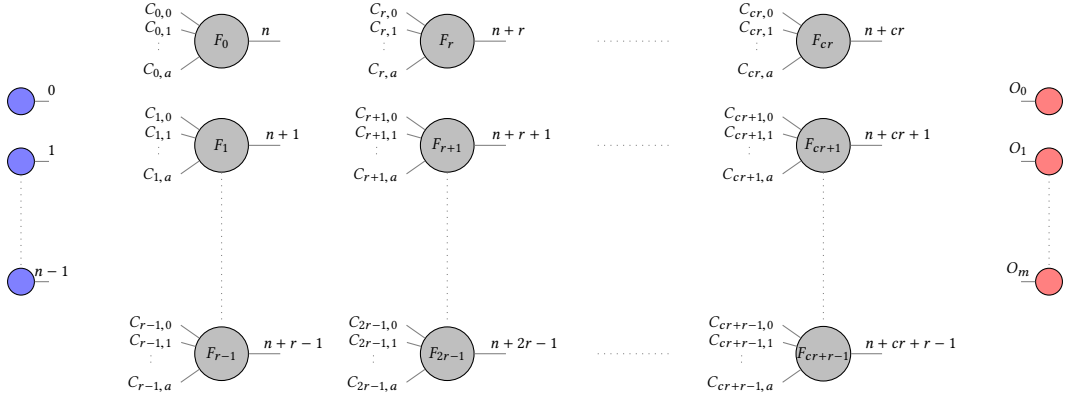
\begin{figure*}[hbtp]
\centering
\def\nodesize{29pt}
\def\layersep{2.5cm}
\def\nodesep{\nodesize*1.7}

\begin{tikzpicture}[shorten >=1pt,->,draw=black!50, node distance=\layersep, scale=0.8, every node/.style={scale=0.7}]
    \tikzstyle{cgp_neuron}=[circle,fill=black!25,minimum size=5pt,inner sep=0pt]
    \tikzstyle{cgp_input} =[cgp_neuron, fill=blue!50, minimum size=\nodesize/2, draw=black];
    \tikzstyle{cgp_output}=[cgp_neuron, fill=red!50, minimum size=\nodesize/2, draw=black];
    \tikzstyle{cgp_node}=[cgp_neuron, minimum size=\nodesize, draw=black];


\path[] node[cgp_input] (I0) at (-\layersep/3,-1cm) {};
\draw[-] (I0) -- + (\nodesize/2,0) node [right, above] (IO0) {$0$};

\path[] node[cgp_input] (I1) at (-\layersep/3,-2cm) {};
\draw[-] (I1) -- + (\nodesize/2,0) node [right, above] (IO1) {$1$};

\path[] node[cgp_input] (I3) at (-\layersep/3,-4cm) {};
\draw[-] (I3) -- + (\nodesize/2,0) node [right, above] (IO2) {$n-1$};

\draw[dotted, -]  (I1) -- (I3);


\path[] node[cgp_output] (O0) at (\layersep*6.5,-1cm) {};
\draw[-] (O0) -- + (-\nodesize/2,0) node [right, above] (OO0) {$O_0$};

\path[] node[cgp_output] (O1) at (\layersep*6.5,-2cm) {};
\draw[-] (O1) -- + (-\nodesize/2,0) node [right, above] (OO1) {$O_1$};

\path[] node[cgp_output] (O3) at (\layersep*6.5,-4cm) {};
\draw[-] (O3) -- + (-\nodesize/2,0) node [right, above] (OO2) {$O_m$};

\draw[dotted, -]  (O1) -- (O3);

    \foreach \name / \y in {0} {
        \path[]
            node[cgp_node] (N1-\name) at (\layersep,-\nodesep*\y) {$F_{\y}$};
        \draw[-] (N1-\name) -- + (\nodesize,0) node [midway, above] (O1-\name) {$n$};
        \draw[-] (N1-\name) -- + (-\nodesize*0.7,0.5) node [left] (C00-\name) {$C_{\y,0}$};
        \draw[-] (N1-\name) -- + (-\nodesize*0.7,0.2) node [left] (C01-\name) {$C_{\y,1}$};
        \draw[-] (N1-\name) -- + (-\nodesize*0.7,-0.5) node [left] (C0a-\name) {$C_{\y, a}$};
        \draw[dotted, -]  (C01-\name) -- (C0a-\name);
    }
    \foreach \name / \y in {1} {
        \path[]
            node[cgp_node] (N1-\name) at (\layersep,-\nodesep*\y) {$F_{\y}$};
        \draw[-] (N1-\name) -- + (\nodesize,0) node [right, above] (O1-\name) {$n + \y$};
        \draw[-] (N1-\name) -- + (-\nodesize*0.7,0.5) node [left] (C00-\name) {$C_{\y,0}$};
        \draw[-] (N1-\name) -- + (-\nodesize*0.7,0.2) node [left] (C01-\name) {$C_{\y,1}$};
        \draw[-] (N1-\name) -- + (-\nodesize*0.7,-0.5) node [left] (C0a-\name) {$C_{\y, a}$};
        \draw[dotted, -]  (C01-\name) -- (C0a-\name);
    }
    \foreach \name / \y in {3} {
        \path[]
            node[cgp_node] (N1-\name) at (\layersep,-\nodesep*\y) {$F_{r-1}$};
        \draw[-] (N1-\name) -- + (\nodesize,0)  node [right, above] (O1-\name) {$\hskip0.4cm n + {r-1}$};
        \draw[-] (N1-\name) -- + (-\nodesize*0.7,0.5) node [left] (C00r) {$C_{r-1,0}$};
        \draw[-] (N1-\name) -- + (-\nodesize*0.7,0.2) node [left] (C01r) {$C_{r-1,1}$};
        \draw[-] (N1-\name) -- + (-\nodesize*0.7,-0.5) node [left] (C0ar) {$C_{r-1, a}$};
        \draw[dotted, -]  (C01r) -- (C0ar);
    }
    \draw[dotted, -]  (N1-1) -- (N1-3);


    \foreach \name / \y in {0} {
        \path[]
            node[cgp_node] (N2-\name) at (2.5*\layersep,-\nodesep*\y) {$F_{r}$};
        \draw[-] (N2-\name) -- + (\nodesize,0)  node [right, above] (O1-\name) {$n + {r}$};
        \draw[-] (N2-\name) -- + (-\nodesize*0.7,0.5) node [left] (Cr0-\name) {$C_{{r},0}$};
        \draw[-] (N2-\name) -- + (-\nodesize*0.7,0.2) node [left] (Cr1-\name) {$C_{{r},1}$};
        \draw[-] (N2-\name) -- + (-\nodesize*0.7,-0.5) node [left] (Cra-\name) {$C_{{r}, a}$};
        \draw[dotted, -]  (Cr1-\name) -- (Cra-\name);
    }
    \foreach \name / \y in {1} {
        \path[]
            node[cgp_node] (N2-\name) at (2.5*\layersep,-\nodesep*\y) {$F_{r+\y}$};
        \draw[-] (N2-\name) -- + (\nodesize,0)  node [right, above] (O1-\name) {$\hskip0.5cm n + {r+\y}$};
        \draw[-] (N2-\name) -- + (-\nodesize*0.7,0.5) node [left] (Cr0-\name) {$C_{{r+\y},0}$};
        \draw[-] (N2-\name) -- + (-\nodesize*0.7,0.2) node [left] (Cr1-\name) {$C_{{r+\y},1}$};
        \draw[-] (N2-\name) -- + (-\nodesize*0.7,-0.5) node [left] (Cra-\name) {$C_{{r+\y}, a}$};
        \draw[dotted, -]  (Cr1-\name) -- (Cra-\name);
    }
    \foreach \name / \y in {3} {
        \path[]
            node[cgp_node] (N2-\name) at (2.5*\layersep,-\nodesep*\y) {$F_{2r-1}$};
        \draw[-] (N2-\name) -- + (\nodesize,0)  node [right, above] (O1-\name) {$\hskip0.6cm n+2r-1$};
        \draw[-] (N2-\name) -- + (-\nodesize*0.7,0.5) node [left] (C00r) {$C_{2r-1,0}$};
        \draw[-] (N2-\name) -- + (-\nodesize*0.7,0.2) node [left] (C01r) {$C_{2r-1,1}$};
        \draw[-] (N2-\name) -- + (-\nodesize*0.7,-0.5) node [left] (C0ar) {$C_{2r-1, a}$};
        \draw[dotted, -]  (C01r) -- (C0ar);
    }
    \draw[dotted, -]  (N2-1) -- (N2-3);

\foreach \name / \y in {0,1,3} {
    \coordinate (D1-\name) at (3.5*\layersep,-\nodesep*\y);
    \coordinate (D2-\name) at (4*\layersep,-\nodesep*\y);
    \draw[dotted, -]  (D1-\name) -- (D2-\name);
}


    \foreach \name / \y in {0} {
        \path[]
            node[cgp_node] (N3-\name) at (5*\layersep,-\nodesep*\y) {$F_{cr}$};
        \draw[-] (N3-\name) -- + (\nodesize,0)  node [right, above] (O1-\name) {$n + {cr}$};
        \draw[-] (N3-\name) -- + (-\nodesize*0.7,0.5) node [left] (Cr0-\name) {$C_{{cr},0}$};
        \draw[-] (N3-\name) -- + (-\nodesize*0.7,0.2) node [left] (Cr1-\name) {$C_{{cr},1}$};
        \draw[-] (N3-\name) -- + (-\nodesize*0.7,-0.5) node [left] (Cra-\name) {$C_{{cr}, a}$};
        \draw[dotted, -]  (Cr1-\name) -- (Cra-\name);
    }
    \foreach \name / \y in {1} {
        \path[]
            node[cgp_node] (N3-\name) at (5*\layersep,-\nodesep*\y) {$F_{cr+\y}$};
        \draw[-] (N3-\name) -- + (\nodesize,0)  node [right, above] (O1-\name) {$\hskip0.5cm n + {cr+\y}$};
        \draw[-] (N3-\name) -- + (-\nodesize*0.7,0.5) node [left] (Cr0-\name) {$C_{{cr+\y},0}$};
        \draw[-] (N3-\name) -- + (-\nodesize*0.7,0.2) node [left] (Cr1-\name) {$C_{{cr+\y},1}$};
        \draw[-] (N3-\name) -- + (-\nodesize*0.7,-0.5) node [left] (Cra-\name) {$C_{{cr+\y}, a}$};
        \draw[dotted, -]  (Cr1-\name) -- (Cra-\name);
    }
    \foreach \name / \y in {3} {
        \path[]
            node[cgp_node] (N3-\name) at (5*\layersep,-\nodesep*\y) {$F_{cr+r-1}$};
        \draw[-] (N3-\name) -- + (\nodesize,0)  node [right, above] (O1-\name) {$\hskip1cm n+cr+r-1$};
        \draw[-] (N3-\name) -- + (-\nodesize*0.7,0.5) node [left] (C00r) {$C_{{cr+r-1},0}$};
        \draw[-] (N3-\name) -- + (-\nodesize*0.7,0.2) node [left] (C01r) {$C_{{cr+r-1},1}$};
        \draw[-] (N3-\name) -- + (-\nodesize*0.7,-0.5) node [left] (C0ar) {$C_{{cr+r-1}, a}$};
        \draw[dotted, -]  (C01r) -- (C0ar);
    }
    \draw[dotted, -]  (N3-1) -- (N3-3);

\end{tikzpicture}
\caption{Most widely used form of Cartesian genetic programming, as described by \cite{miller2011cartesian}. \label{fig:cgp}}
\end{figure*}

\subsection{Cartesian Genetic Programming}
\label{subsec:genprog}

In its original form, CGP~\cite{miller2011cartesian} has been deployed to various applications, including the evolution of robotic controllers~\cite{Harding2005EvolutionOR}, digital filters~\cite{Miller1999Evolution}, computational art~\cite{Ashmore2003EvolutionaryAW} and large scale digital circuits~\cite{Vasicek2015Cartesian}. The idea to use the CGP-encoding to represent neural networks goes back to works of Turner and Miller~\cite{turner2013cartesian} and Khan et al.~\cite{khan2013fast}, who coined the term CGPANN. In these works, the network parameters (mainly weights as no biases were introduced) are evolved by genetic operators, and the developed techniques are thus applied to tasks where gradient information is not available (e.g. reinforcement learning). In contrast, our work will make explicit use of the gradient information for adapting weights and node biases, effectively creating a memetic algorithm~\cite{Moscato2004Memetic}. There exists some work on the exploitation of low-order differentials to learn parameters for genetic programs in general~\cite{topchy2001faster, emigdio2015local} but the application of gradient descent to CGPANNs is widely unexplored.

A notable exception is the recent work of Suganuma et al.~\cite{suganuma2017genetic}, who deployed CGP to encode the interconnections among functional blocks of a convolutional neural network. In~\cite{suganuma2017genetic}, the nodes of the CGP represent highly functional modules such as convolutional blocks, tensor concatenation and similar operations. The resulting convolutional neural networks are then trained by stochastic gradient descent. In contrast, Our approach works directly on the fundamental units of computation, i.e. neurons, activation functions and their connectome.

\subsection{Neural Network Architecture Search}
\label{subsec:nnsearch}

There is great interest to automate the design process of neural networks, since finding the best performing topologies by human experts is often viewed as a laborious and tedious process. Some recent approaches deploy Bayessian optimization~\cite{Snoek2012Practical} or reinforcement learning~\cite{zoph2016neural, baker2016designing} to discover architectures that can rival human designs for large image classification problems like CIFAR-10 or CIFAR-100. However, automated architecture design often comes with heavy ressource requirements and many works are dedicated to mitigate this issue~\cite{cai2018efficient, fang2019eat, real2017large}. 

One way to perform architecture search are metaheuristics and neuro-evolution, which have been studied since decades and remain a profilific area of research~\cite{Ojha2017MetaheuristicDO}. Most notably, NEAT~\cite{stanley2002evolving} and its variations~\cite{stanley2009hypercube, miikkulainen2019evolving} have been investigated as methods to grow network structures while simultaneously evolving their corresponding weights. The approach by Han et al.~\cite{han2015learning} is almost orthogonal, as it deploys an effective pruning strategy to learns weights and topology purely from gradients. Our approach takes aspects of both: weights are learned from gradients while network topologies are gradually improved by evolution.

We focus on small-scale regression problems and optimize our topologies for efficient training as a means to combat the exploding resource requirements. In this sense, our approach is related to the concept of meta-learning~\cite{Vilalta2002}; the ability to ``learn how to learn'' by exploiting meta-knowledge and adapting to the learning task at hand. The idea to evolve effective learning systems in form of neural networks during their training has recently been surveyed by Soltoggio et al.~\cite{soltoggio2018born}, who coin the term ``EPANN'' (Evolved Plastic Artificial Neural Network). However, to the best of our knowledge, our work is the first to analyze plasticity in neural networks represented as CGPs.

\section{Differentiable Cartesian Genetic Programming}
\label{sec:dcgp}
This section outlines our design of a neural network as a CGP and explains how it can be trained efficiently.

\subsection{Definition of a dCGPANN}
\label{subsec:dcgpann}

A Cartesian genetic program \citep{miller2011cartesian}, in the widely used form depicted in Figure \ref{fig:cgp}, is defined by the number of inputs $n$, the number of outputs $m$, the number of rows $r$, the number of columns $c$, the levels-back $l$, the arity $a$ of its kernels (non-linearities) and the set of possible kernel functions. With reference to the picture, each of the $n + rc$ nodes in a CGP is thus assigned a unique id and the vector of integers:
$$
\mathbf x_I = [F_0, C_{0,0}, C_{0,1}, ...,  C_{0, a}, F_1, C_{1,0}, ....., O_1, O_2, ..., O_m]
$$
defines entirely the value of the terminal nodes. Indicating the numerical value of the output of the generic CGP node having id $i$ with the symbol $N_i$, we formally have that:
$$
N_i = F_i\left(N_{C_{i,0}}, N_{C_{i,1}}, ..., N_{C_{i,a}}\right)
$$
In other words, each node outputs the value of its kernel -- or non linearity, to adopt a terminology more used in ANN research -- computed using as inputs the connected nodes. 

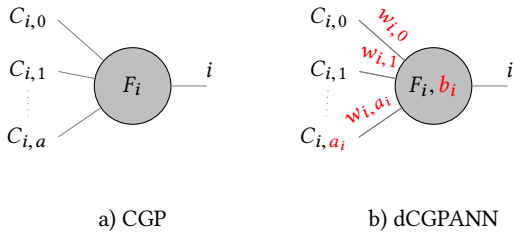
\begin{figure}[btp]
\centering
\def\nodesize{29pt}
\def\layersep{1cm}
\def\nodesep{\nodesize*1.7}

\begin{tikzpicture}[shorten >=1pt,->,draw=black!50, node distance=\layersep]
    \tikzstyle{cgp_neuron}=[circle,fill=black!25,minimum size=5pt,inner sep=0pt]
    \tikzstyle{cgp_input} =[cgp_neuron, fill=blue!50, minimum size=\nodesize/2, draw=black];
    \tikzstyle{cgp_output}=[cgp_neuron, fill=red!50, minimum size=\nodesize/2, draw=black];
    \tikzstyle{cgp_node}=[cgp_neuron, minimum size=\nodesize, draw=black];

\path[]
    node[cgp_node] (N1) at (\layersep,0) {$F_{i}$};
\draw[-] (N1) -- + (\nodesize,0) node [right, above] (O1) {$i$};
\draw[-] (N1) -- + (-\nodesize,0.9) node [left] (C00) {$C_{i,0}$};
\draw[-] (N1) -- + (-\nodesize,0.2) node [left] (C01) {$C_{i,1}$};
\draw[-] (N1) -- + (-\nodesize,-0.7) node [left] (C0a) {$C_{i, a}$};
\draw[dotted, -]  (C01) -- (C0a);
\node (CGP) [below=of N1] {a) CGP};

\path[]
    node[cgp_node] (N2) at (5*\layersep,0) {$F_{i}, \textcolor{red}{b_i}$};
\draw[-] (N2) -- + (\nodesize,0) node [right, above] (O1) {$i$};
\draw[-] (N2) -- + (-\nodesize,0.9) node [left] (C10) {$C_{i,0}$};
\draw[-] (N2) -- + (-\nodesize,0.9) node [midway, sloped, above] (C11) {$\textcolor{red}{w_{i,0}}$};
\draw[-] (N2) -- + (-\nodesize,0.2) node [left] (C20) {$C_{i,1}$};
\draw[-] (N2) -- + (-\nodesize,0.2) node [midway, sloped, above] (C21) {$\textcolor{red}{w_{i,1}}$};
\draw[-] (N2) -- + (-\nodesize,-0.7) node [left] (C2a) {$C_{i, \textcolor{red}{a_i}}$};
\draw[-] (N2) -- + (-\nodesize,-0.7) node [midway, sloped, above] (C2a1) {$\textcolor{red}{w_{i,a_i}}$};
\draw[dotted, -]  (C20) -- (C2a);
\node (dCGPANN) [below=of N2] {b) dCGPANN};

\end{tikzpicture}
\caption{Differences between the $i$-th node in a CGP expression and in a dCGPANN expression. \label{fig:node_diff}}
\end{figure}

We modify the standard CGP node adding a weight $w$ for each connection $C$, a bias $b$ for each function $F$ and a different arity $a$ for each node. We also change the definition of $N_i$ to:
\begin{equation}
\label{eq:node_value}
N_i = F_i\left(\sum_{j=0}^{a_i} w_{i,j} N_{C_{i,j}} + b_j\right)
\end{equation}
forcing the non linearities to act on the biased sum of their weighted inputs. Note that, by doing so, changing the function arity corresponds to adding or removing terms from the sum. The difference between a standard CGP node and a dCGP is depicted in Figure \ref{fig:node_diff}. 

We define $\mathbf x_R$ as the vector of real numbers:
$$
\mathbf x_R = \theta = [b_0, w_{0,0}, w_{0,1}, ...,  w_{0, a_0}, b_1, w_{1,0}, w_{1,1}, ...,  w_{1, a_1}, ...]
$$
The two vectors $\mathbf x_I$ and  $\mathbf x_R$ form the chromosome of our dCGPANN and suffice for the evaluation of the terminal values $O_i, i=1..m$. Note that we also have introduced the symbol $\theta$ to indicate $\mathbf x_R$ as the network parameters are often indicated with this symbol in machine learning related literature.

\subsection{Training of dCGPANNs}
\label{subsec:train}
Training dCGPANNs is more complex than training ANNs as both the continuous and the integer part of the chromosome affects the loss $\ell$ and have thus to be learned.
The obvious advantage is that, when successful, in a dCGPANN the learning will not only result in network parameters $\theta$ adapted to the task, but also in a network topology adapted to the task, including its size, activation functions and connections.

Consider a regression task where some loss $\ell$ is defined. Assume the integer part of the chromosome $\mathbf x_I$ is known and fixed. It is possible to compute efficiently the loss gradient $\nabla_{\mathbf x_R} \ell  = \frac{d\ell}{d\mathbf x_R}$ implementing a backward automated differentiation scheme over the computational graph defined by the dCGPANN. This is not a straight forward task as the computational graph of a dCGPANN is much more intricate than that of a simple ANN, but it is attainable and eventually leads to the possibility of computing the gradient with a little computational overhead independent of $\theta$ (constant complexity). Note also that if $\mathbf x_I$ is changed, a new backward automated differentiation scheme needs to be derived for the changed computational graph. 

Once the loss gradient is computed, the classical gradient descent rule can be used to update $\theta$:

\begin{equation}
\label{eq:update_rule}
\theta_{i+1} = \theta_i - lr * \nabla_{\mathbf \theta} \ell
\end{equation}

\begin{figure}[btp]
\centering
\includegraphics[width=0.45\textwidth]{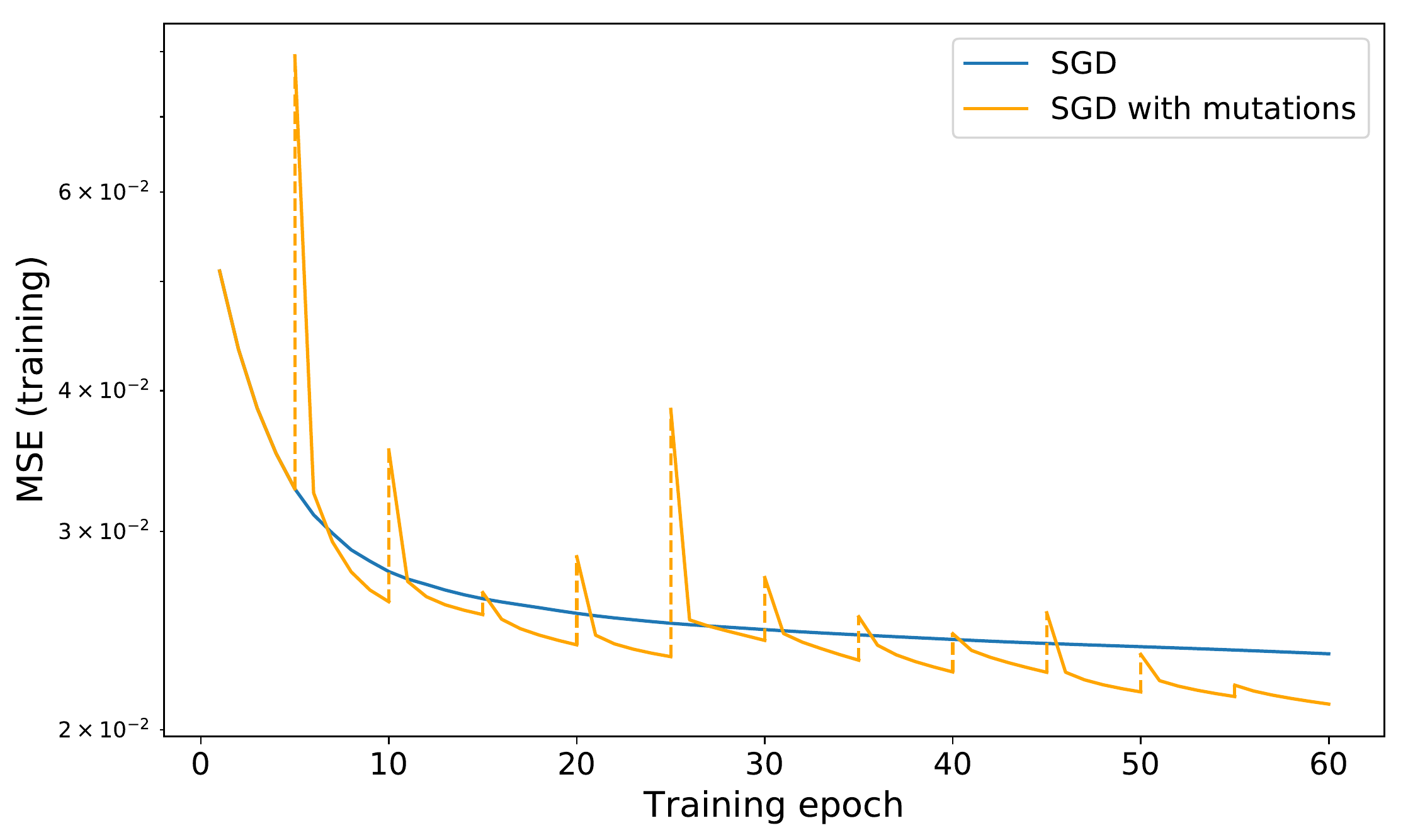}
\caption{Loss during a generic regression task: SGD is compared with SGD perturbed, each 5 epochs, by a random (cumulative) mutation. }
\label{fig:combo}
\end{figure}
\noindent 
where $lr$ is used to indicate the learning rate (i.e. a trust region for the first order Taylor expansion of the loss). Epoch after epoch and batch after batch, the above update rule can be used to assemble a stochastic gradient descent algorithm during which -- between epochs --  the integer part of the chromosome $\mathbf x_I$ may also change subject to different operators, typically evolutionary ones such as mutation and crossover.

Regardless of the actual details of such a learning scheme, it is of interest to study how the SGD update rule gets disturbed by applying an evolutionary operator to $\mathbf x_I$. In Figure~\ref{fig:combo} we show, for a generic regression task, the trend of the training error when the update rule in Eq.(\ref{eq:update_rule}) is disturbed, each fixed number of epochs, by mutating some active gene in $\mathbf x_I$. The details are not important here as the observed behaviour is generally reproducible with different tasks, networks, optimizers and evolutionary operators, but it is, instead, important to note a few facts: firstly, mutations immediately degrade the loss. This is expected since the parameters are learned on a different topology before the mutation occurs. However, the loss recovers relatively fast after a few more batches of data pass through SGD. Secondly, mutations can be ultimately beneficial as after already a few epochs the error level can be lower than what one would obtain if no mutations had occurred. In other words, we can claim that genetic operators applied on $\mathbf x_I$ to a stochastic gradient descent learning process working on $\mathbf x_R$ are mostly detrimental in the short term, but may offer advantages in the long term.

With this in mind, it is clear that the possibilities to assemble one single learning scheme mixing two powerful tools -- evolutionary operators and the SGD update rule -- are quite vast and their success will be determined by a clever application of the genetic operators and by protecting them for some batches/epochs before either selection or removal is applied. Note how this scheme is, essentially, a memetic approach where the individual learning component corresponds to the SGD learning of $\theta$.
In this work (Section~\ref{subsec:lsmf}) we will propose one easy set-up for a memetic algorithm that was found to deliver most interesting results over the test cases chosen. It makes use of Lamarckian inheritance when selecting good mutations, but it also introduces a purely Darwinian mechanism (the ``Forget step'') that allows for a complete reset of its accumulated experiences.

The implementation details of the dCGPANN and, most importantly, of the backward automated differentiation scheme, have been released as part of the latest release of the open source project on differentiable Cartesian Genetic Programming (dCGP)~\cite{Izzo2017Differentiable}. No SIMD vectorization nor GPU support is available in this release as it will, instead, be the subject of the future developments. The parallelization available so far relies on multithreading and affects the parallel evaluation of the dCGPANN expression and its gradient for data within each batch during one SGD epoch. This led us, for this paper, to experiment mainly with tasks appropriate to small networks where the performances of our code is comparable, timewise, with that of popular deep learning frameworks such as tensorflow and pytorch. Our results have thus to be taken as valid for small scale networks and their scalability to larger sizes remains to be shown and rests upon the development of an efficient, vectorized version of our current code base.
 
\section{Experiments}

We conduct experiments to show how evolution can be applied to dCGPANNs in order to enhance their learning capabilities by continuous improvements on their network topologies. Details about the data, algorithms and our experimental setups are reported in the following.

\label{sec:experiments}

\begin{figure}
\centering
\includegraphics[width=4.2cm]{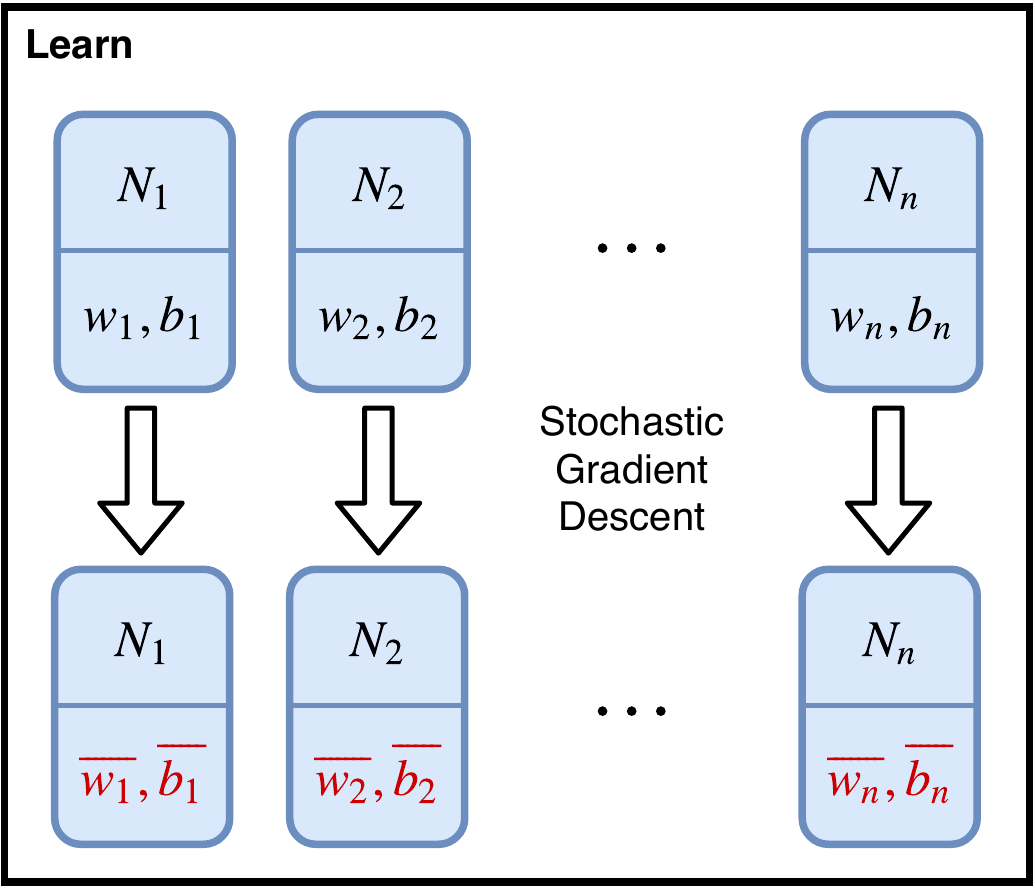}
\includegraphics[width=4.2cm]{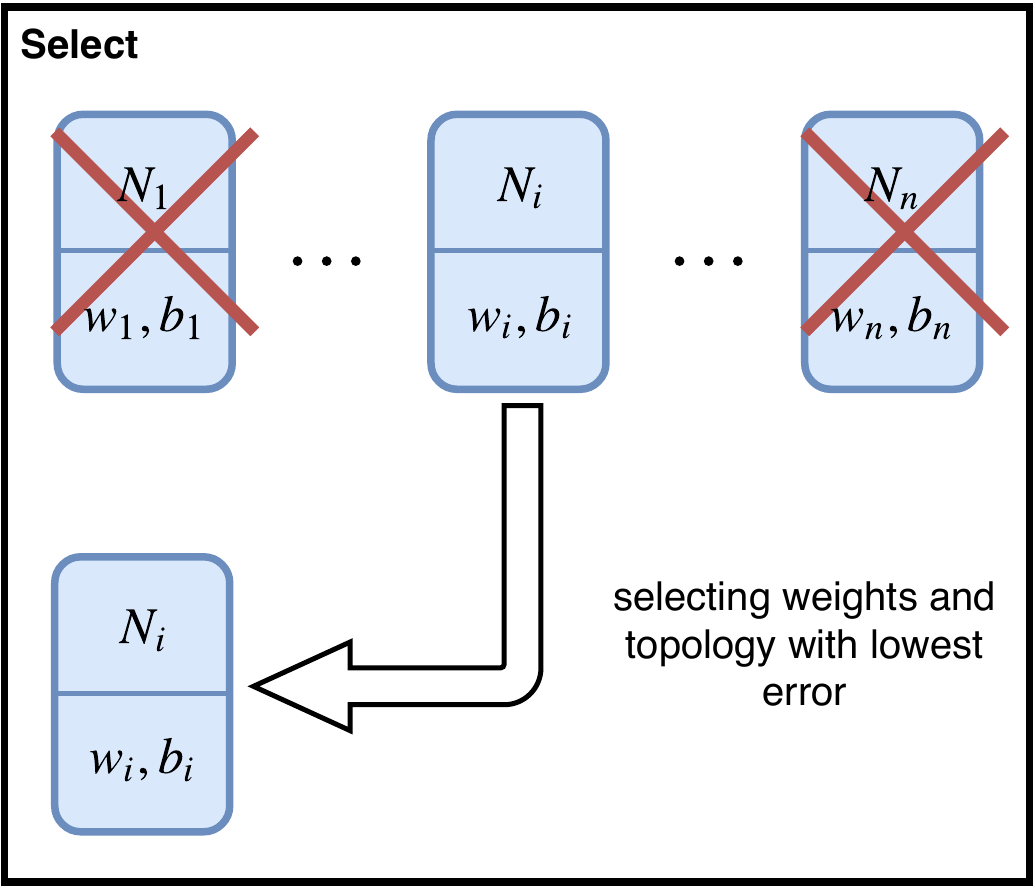}
\includegraphics[width=4.2cm]{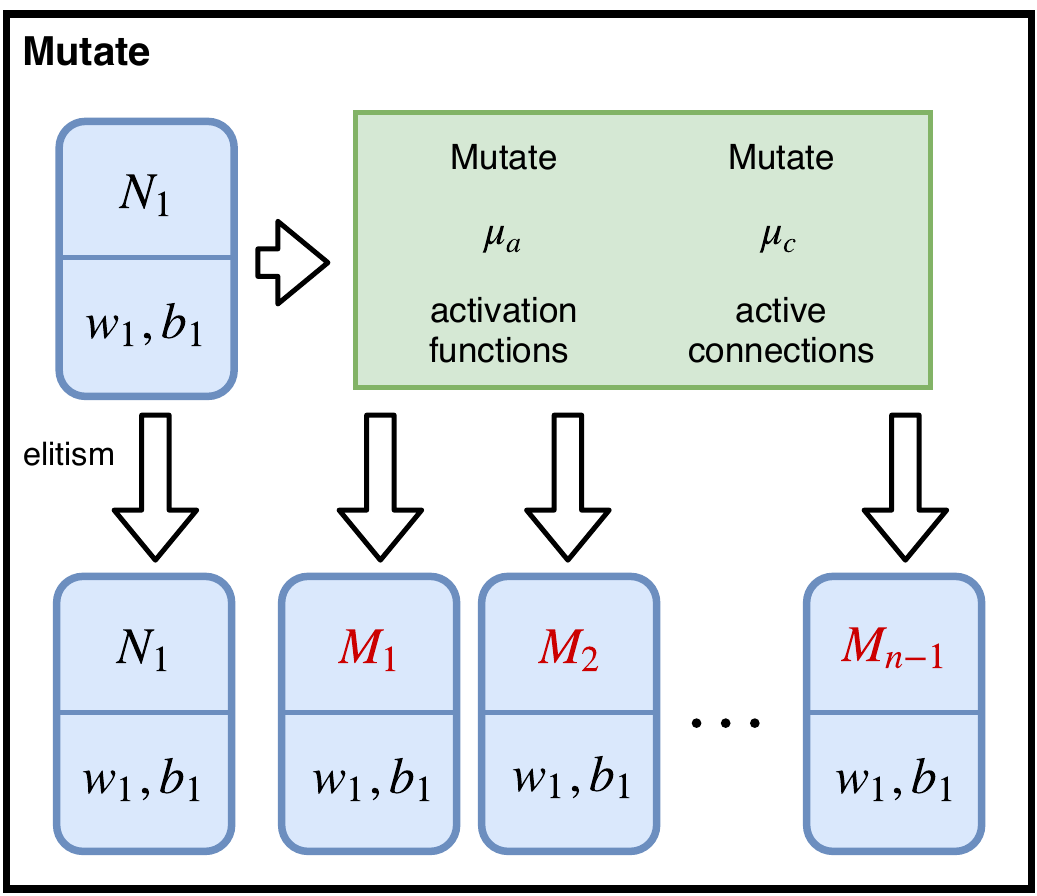}
\includegraphics[width=4.2cm]{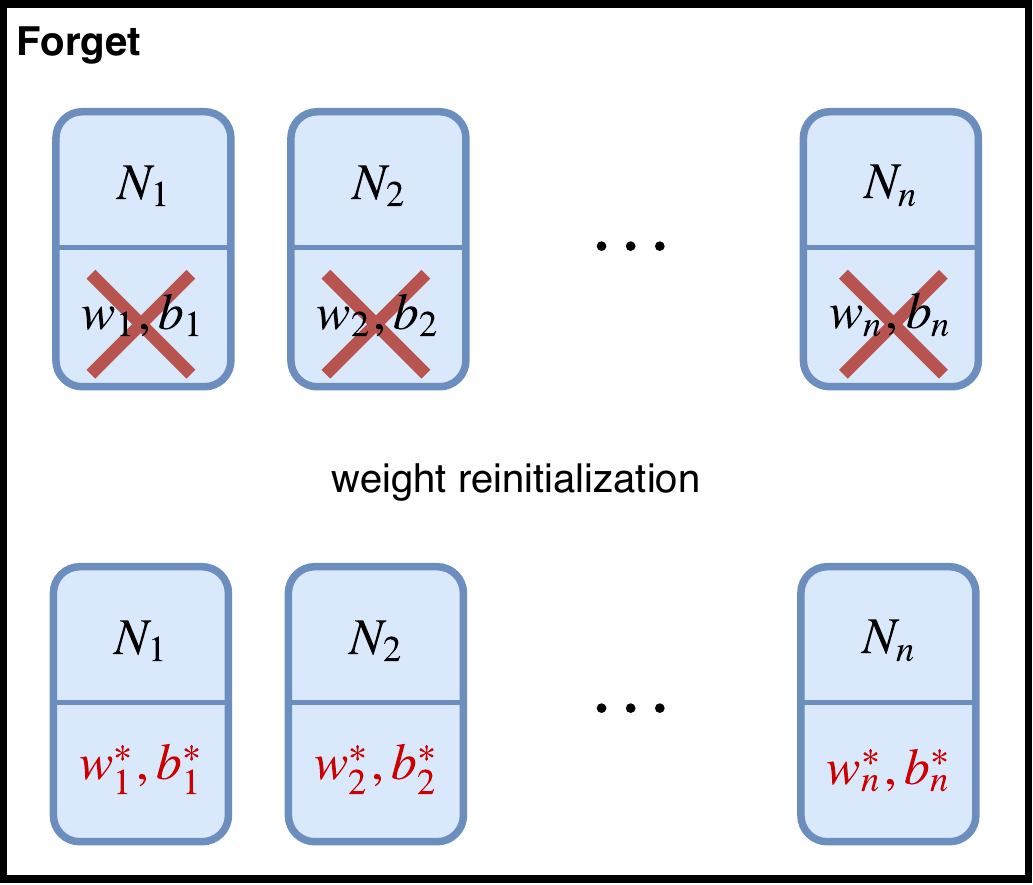}
\caption{Each block shows one of the four operations of the LSMF algorithm. In each block, the top line corresponds to the input population of the step and the bottom line highlights the applied changes (output population).}
\label{fig:alg_schematics}
\end{figure}

\subsection{Datasets}
\label{subsec:datasets}

Our experiments are based on 6 regression problems, which we imported directly from the Penn Machine Learning Benchmarks (PMLB)~\cite{Olson2017PMLB}. We selected these problems for diversity and distinct complexity, while avoiding challenges with too many or too few instances. Table~\ref{tab:datasets} shows the name of the dataset with the corresponding number of instances and features. Each problem is listed on \url{www.openml.org} where meta-data and descriptions can be found.

\begin{table}[btp]
	\begin{tabular}{lcc}
	\hline 
	\textbf{name} & \textbf{\#instances} & \textbf{\#features} \\ 
	\hline 
	197\_cpu\_act & 8192 & 21 \\ 
	225\_puma8NH & 8192 & 8 \\ 
	344\_mv & 40768 & 10 \\ 
	503\_wind & 6574 & 14 \\ 
	564\_fried & 40768 & 10 \\ 
	1201\_BNG\_breastTumor & 116640 & 9 \\ 
	\hline 
	\end{tabular} 
	\caption{Regression problems selected for experiments.}
	\label{tab:datasets}
\end{table}

The features of every dataset are standardized and the single target value minmax-scaled to the unit-interval. Data is split 75/25 for training/testing.

\subsection{The LSMF Algorithm}
\label{subsec:lsmf}

The LSMF algorithm (short for "Learn, Select, Mutate and Forget") works in $J$ iterations, each consisting of $K$ cycles. A cycle represents a short period of learning for a population of $N$ dCGPANNs followed by a mutation of the best performing network. The steps that are executed during one cycle are the following:

\begin{itemize}
\item[1.] \textbf{Learn}\\
For each of the $N$ dCGPANNs, run $C$ epochs of stochastic gradient descent with learning rate $lr$ and mini-batches of size $mb$. This step only changes the $\mathbf x_R$ part of the chromosomes, keeping the topology of the networks $\mathbf x_I$ constant.
\item[2.] \textbf{Select}\\
For each dCGPANN, compute the loss $\ell$ (i.e. training error) and select the best individual from the population (minimum $\ell$), eliminating all others.
\item[3.] \textbf{Mutate}\\
From the selected dCGPANN, create $N-1$ new networks by mutating a fraction of $\mu_a$ active function genes and a fraction of $\mu_c$ active connectivity genes. Elitism ensures that the new population contains the unaltered original next to its $N-1$ mutants. Complementary to Learn, this step operates only on $\mathbf x_I$, but leaves $\mathbf x_R$ unchanged.
\end{itemize}

Each cycle represents the lifelong learning of the dCGPANNs: after alterations to their network topologies, mutants are protected for $C$ epochs during the train step to enable their development during training (see Figure~\ref{fig:combo}). We call this the \emph{cooldown period}.
Eventually, the weights and biases of the dCGPANNs will converge towards a (near-)optimal loss, regardless of their current network topologies. In this situation, it becomes increasingly difficult for mutants to achieve significant improvements (as the loss is close to optimal) and the development of the topologies stalls. To resolve this situation, we execute a fourth step after $K$ cycles, that reinitializes all weights and biases:

\begin{itemize}
\item[4.] \textbf{Forget}\\
For each dCGPANN, reinitialize all weights and biases while maintaining the network topology. Thus, $\mathbf x_R$ is randomized while $\mathbf x_I$ remains unchanged.
\end{itemize}

After the Forget step, a new evolutionary iteration begins, consisting of another $K$ cycles of Learn, Select and Mutate. The result of each evolutionary iteration is a new dCGPANN network topology, which can be extracted from the last performed Select step. Figure~\ref{fig:alg_schematics} illustrates the four different steps inside LSMF and shows how each manipulates the population of dCGPANNs.

\begin{figure}
\centering
\includegraphics[width=0.3\textwidth]{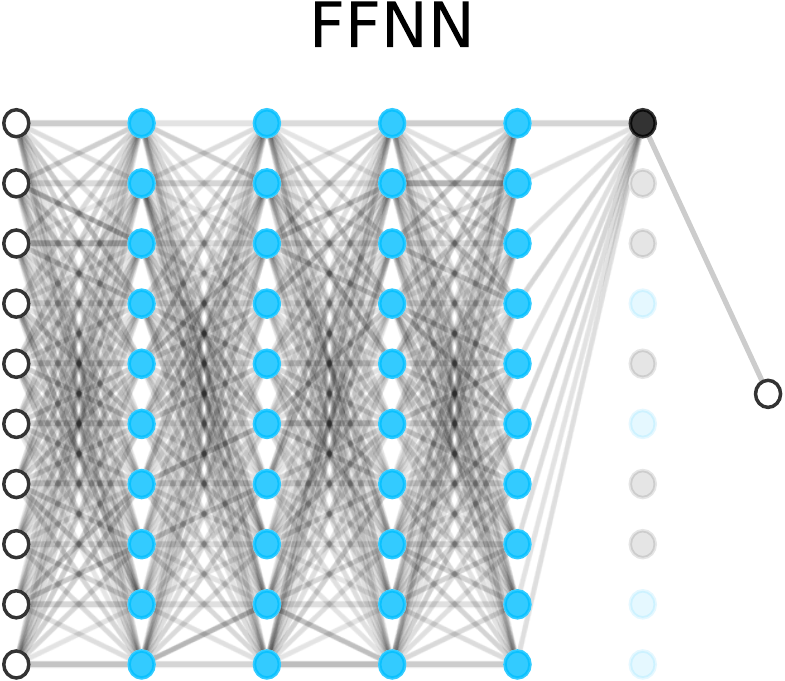}
\includegraphics[width=0.45\textwidth]{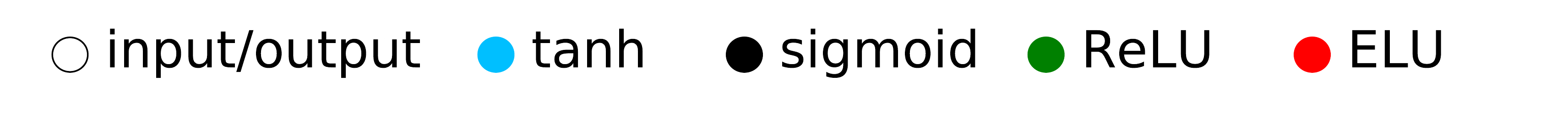}
\caption{Template network for a problem with 10 input features and one output. Number of input nodes will vary with the problem.}
\label{fig:template_network}
\end{figure}

\begin{table}
    \centering
    \begin{tabular}{cc}
    \hline 
    \multicolumn{2}{c}{\textbf{dCGP parameters}} \\
    \hline
    rows $r$ & 10 \\ 
    columns $c$ & 4 \\ 
    arity first column $a_1$ & number of inputs \\ 
    arity other columns $a_2, a_3, a_4$ & 10 \\ 
    levels-back $l$ & 3 \\ 
    set of kernels & \texttt{tanh}, \texttt{sig}, \texttt{ReLU}, \texttt{ELU} \\ 
    \hline
    \multicolumn{2}{c}{\textbf{Evolutionary Parameters}} \\
    \hline
    population $N$ & 100 \\
    mutation rate functions $\mu_a$ & 0.02 \\ 
    mutation rate connections $\mu_c$ & 0.01 \\ 
    cooldown period $C$ & 1 epoch \\ 
    number of cycles per iteration $K$ & 30 \\
    evolutionary iterations $J$ & 10 \\     
    \hline 
    \multicolumn{2}{c}{\textbf{Parameters for Learning}} \\
    \hline
    learning rate $lr$ & 0.01 \\
    batch size $mb$ & 10 \\
    initialization biases & 0 \\
    initialization weights & $\mathcal{N}(0,1)$ \\
    \hline 
    \end{tabular}
    \caption{Set of hyper-parameters for the experiments.}
    \label{tab:hyperparameters}
\end{table}

\subsection{Experimental Setup and Hyperparameters}
\label{subsec:hyperparams}

The LSMF algorithm needs a set of hyperparameters, which can be tuned to adapt to specific problems. Thus, to show the true potential of LSMF, we would need to sample the hyperparameter space for each problem separately and pick the best combination of values, which is beyond the scope of this work. Instead, we want to highlight the broad applicability and general robustness that comes with genetic algorithms like LSMF over multiple regression problems by a reasonable choice of problem-independent fixed values. Table~\ref{tab:hyperparameters} lists our selection of hyperparameters. In the following, we elaborate on some of those choices and provide more details on the experimental setup.

In our experiments, the LSMF algorithm operates on a population of $N = 100$ dCGPANNs, which we initialize as feed-forward neural networks with $4$ hidden layers, consisting of $10$ neurons of activation function \texttt{tanh} each. The single output neuron uses a \texttt{sig} activation function. The set of possible kernels consists of \texttt{tanh}, \texttt{sig}, \texttt{ReLU} and \texttt{ELU}. Figure~\ref{fig:template_network} shows this architecture, which we call the \emph{template network} in the following. Preliminary experiments showed that the template network is able to solve all tasks at hand and provides a good starting point for mutations. Note, that while in the template network only neighboring layers are connected, mutations can cause rewirings to \emph{skip} some of its layers. In particular, we set the levels-back parameter of its dCGPANN representation to $3$, allowing a connection to bypass at most two layers.

Because our population size is $N = 100$, each Mutate step will generate $99$ mutants from the selected best dCGPANN. Each mutation changes 1\% of the connection genes and 2\% of the activation function genes uniformly at random within their respective bounds. Given the fixed structure of our template network, this results in $5$ to $7$ mutated connections (depending on the number of inputs) and $1$ mutated activation function during each Mutate step for the upcoming experiments.

Although all template networks have the same topology, their initial weights and biases are different. In particular, weights are drawn from the standard normal distribution while biases are initialized with zero. Whenever stochastic gradient descent is applied, this happens with a fixed learning rate of $0.1$ and a mini-batch size of $10$. 
The cooldown period $C$ (i.e. the time mutants are trained before selection) is set to $1$ epoch, i.e. a full pass over the training data. For large datasets with millions of instances, a whole epoch might be unnecessarily long, but we found this value to work out for the small-scale data that we analyze.

We execute $K = 30$ cycles of Learn - Select - Mutate for each evolutionary iterations. In total, we perform $J = 10$ evolutionary iterations and track the development of the dCGPANN topology at the end of each.

\section{Results}
\label{sec:results}

\begin{table*}
	\begin{tabular}{lccccc}
	\hline 
	\textbf{name} & 
	\textbf{\#weights} &  
	\Centerstack{\textbf{\#active} \\ \textbf{weights}} & 
	\Centerstack{\textbf{\#skip} \\ \textbf{connections}} & 
	\Centerstack{\textbf{\#duplicate} \\ \textbf{connections}} & 
	\Centerstack{\textbf{compression} \\ \textbf{ratio}} \\
	\hline 
197\_cpu\_act & 520 & 480 & 78 & 98 & 0.734615 \\
225\_puma8NH & 390 & 330 & 90 & 97 & 0.597436 \\
344\_mv & 410 & 370 & 85 & 95 & 0.670732 \\
503\_wind & 450 & 370 & 132 & 108 & 0.582222 \\
564\_fried & 410 & 370 & 81 & 91 & 0.680488 \\
1201\_BNG\_breastTumor & 400 & 350 & 127 & 98 & 0.630000 \\
	\hline
	\end{tabular} 
	\caption{Summary on different connection types in the best evolved dCGPANNs. The column "\#weights" describes the maximum number of weights as used by the initial feed forward neural network. The compression ratio describes by what fraction this number can be reduced in the evolved topology because of either inactive or duplicate connections.}
	\label{tab:compression_skip}
\end{table*}

\begin{table*}[btp]
	\begin{tabular}{lccc}
	\hline 
	\textbf{name} & 
	\Centerstack{\textbf{evolved dCGPANN} \\ \textbf{training error}} & 
	\Centerstack{\textbf{evolved dCGPANN} \\ \textbf{test error}} & 
	\Centerstack{\textbf{random dCGPANN} \\ \textbf{test error}} \\
	\hline 
197\_cpu\_act & 1.851e-03 ($\pm$3.925e-04) & 2.071e-03 ($\pm$4.599e-04) & 2.921e-03 ($\pm$1.748e-03) \\
225\_puma8NH & 2.104e-02 ($\pm$1.428e-03) & 2.031e-02 ($\pm$1.392e-03) & 2.364e-02 ($\pm$3.420e-03) \\
344\_mv & 3.372e-04 ($\pm$1.571e-04) & 3.657e-04 ($\pm$1.555e-04) & 5.655e-04 ($\pm$2.457e-04) \\
503\_wind & 6.799e-03 ($\pm$4.366e-04) & 7.945e-03 ($\pm$4.631e-04) & 8.690e-03 ($\pm$8.353e-04) \\
564\_fried & 2.176e-03 ($\pm$5.324e-04) & 2.878e-03 ($\pm$6.355e-04) & 4.181e-03 ($\pm$1.622e-03) \\
1201\_BNG\_breastTumor & 2.016e-02 ($\pm$2.348e-04) & 2.345e-02 ($\pm$5.472e-04) & 2.436e-02 ($\pm$7.596e-03) \\	
	\hline
	\end{tabular} 
	\caption{Numerical values for mean and standard deviation of training and test error for random and evolved dCGPANNs.}
	\label{tab:funny_numbers}
\end{table*}

\begin{figure*}[tbp]
\centering
\includegraphics[width=1.4cm]{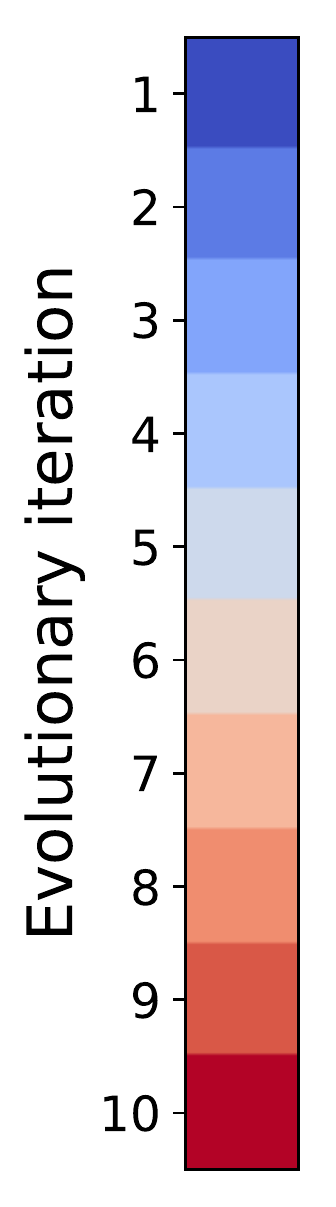}
\includegraphics[height=4.8cm]{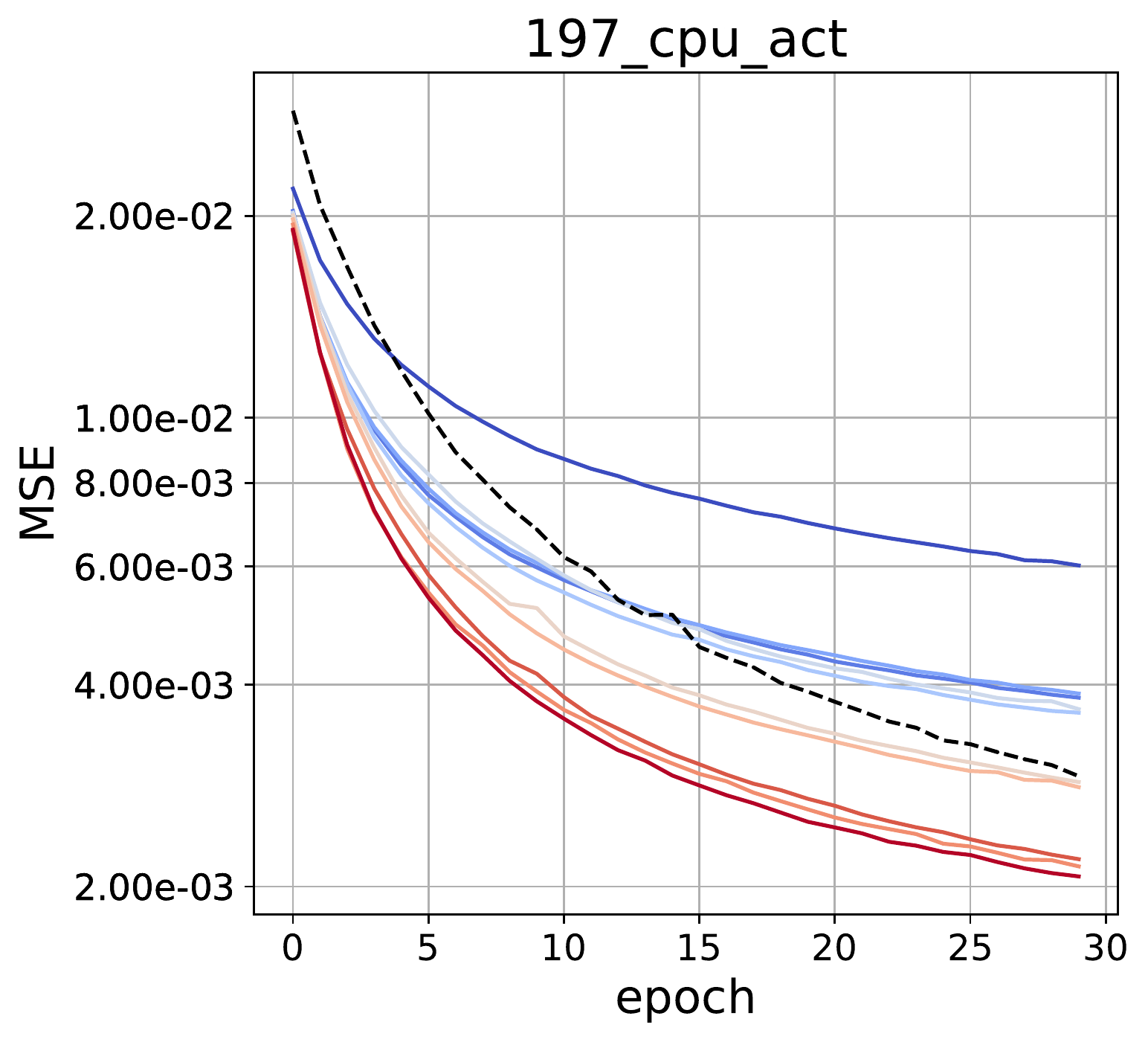}
\includegraphics[height=4.8cm]{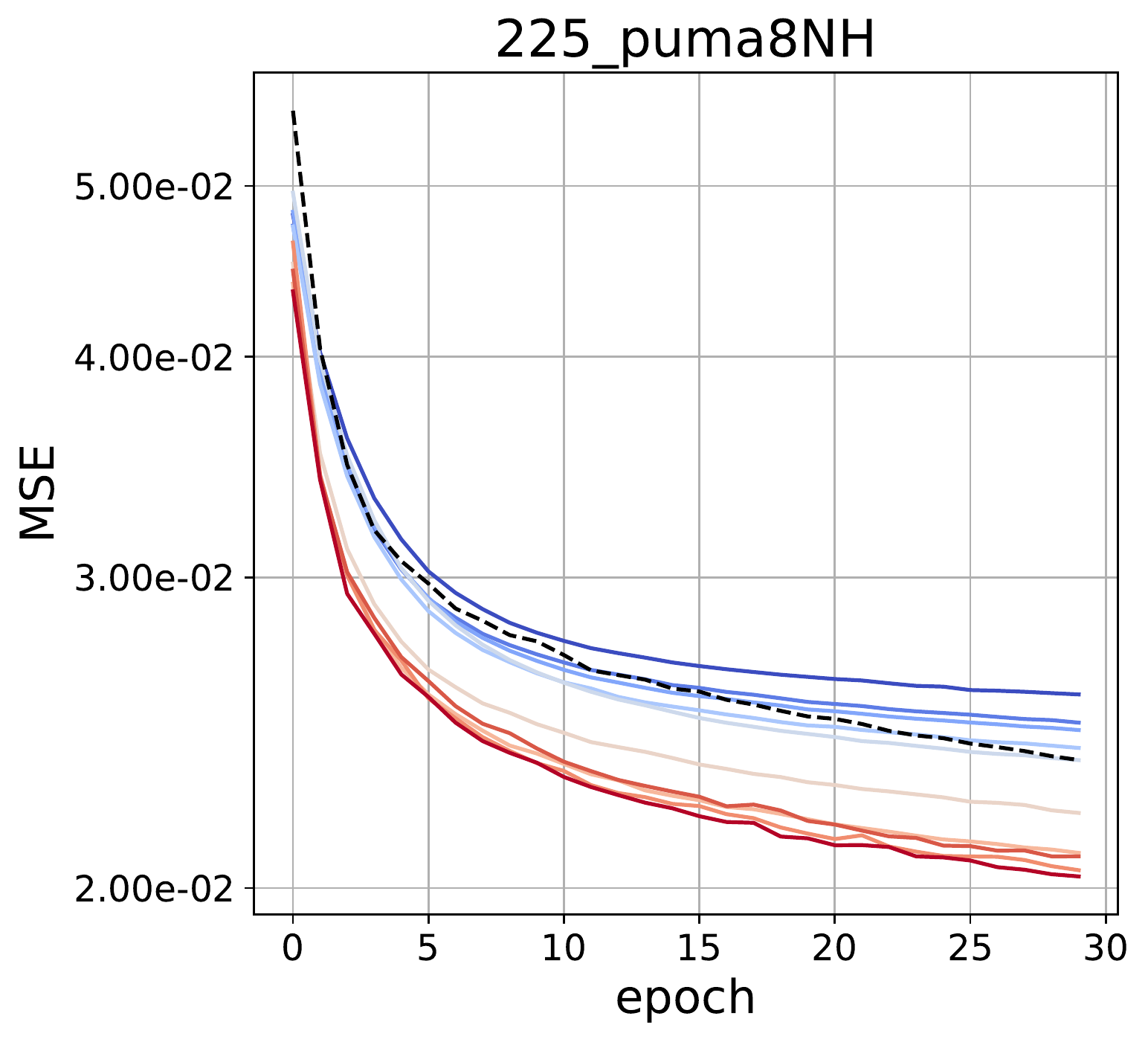}
\includegraphics[height=4.8cm]{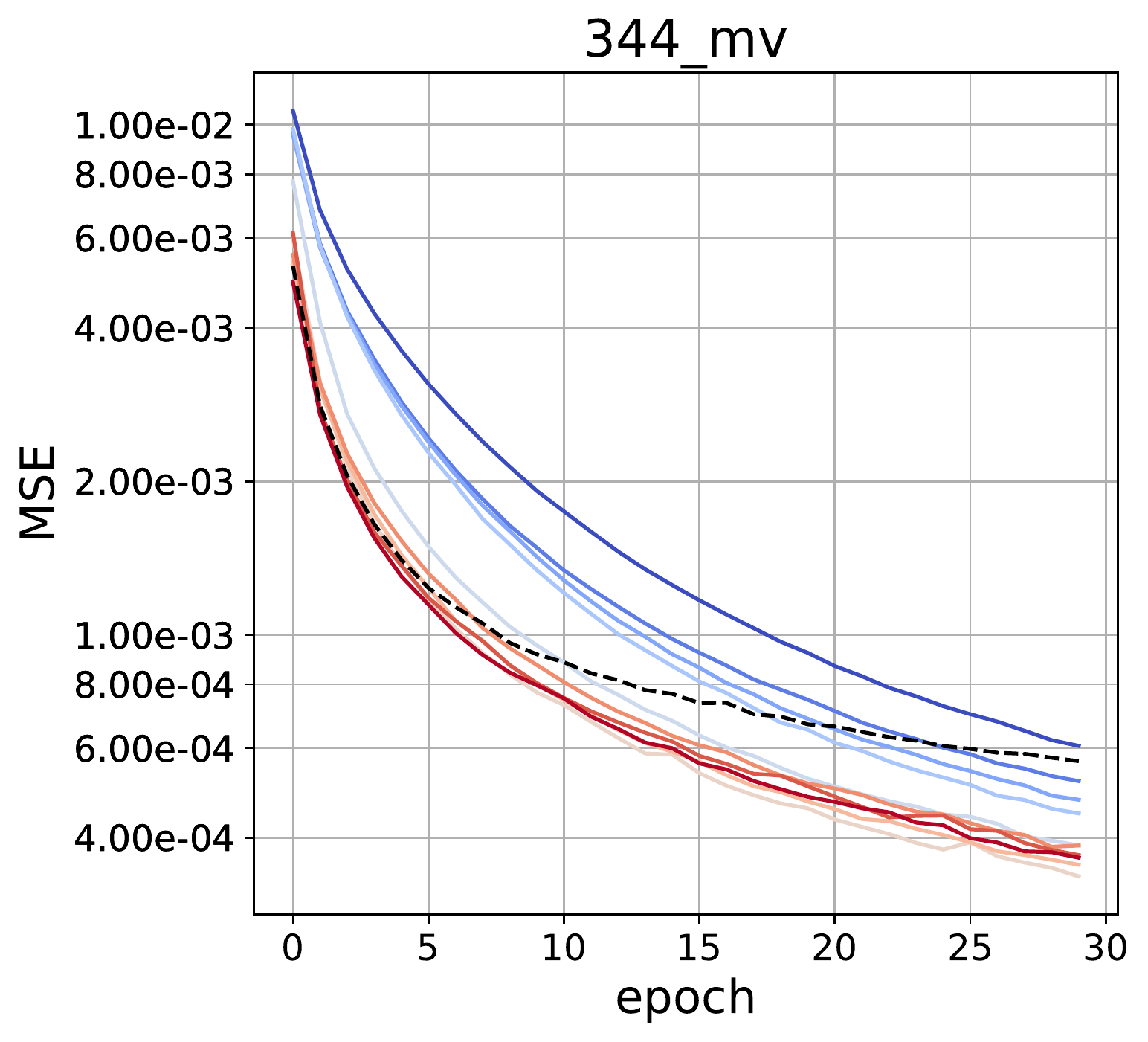}

\hspace*{1.4cm}
\includegraphics[height=4.8cm]{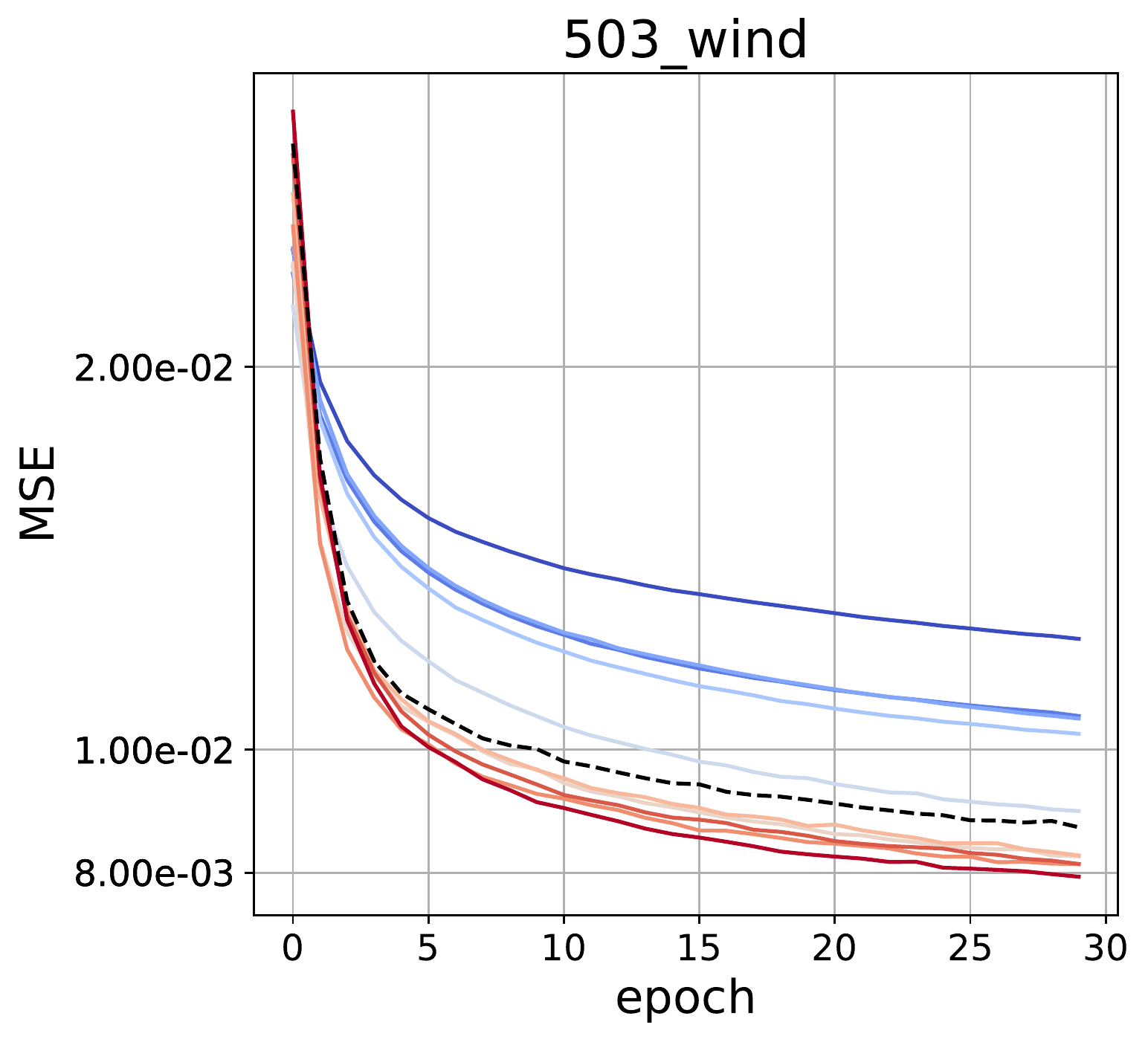}
\includegraphics[height=4.8cm]{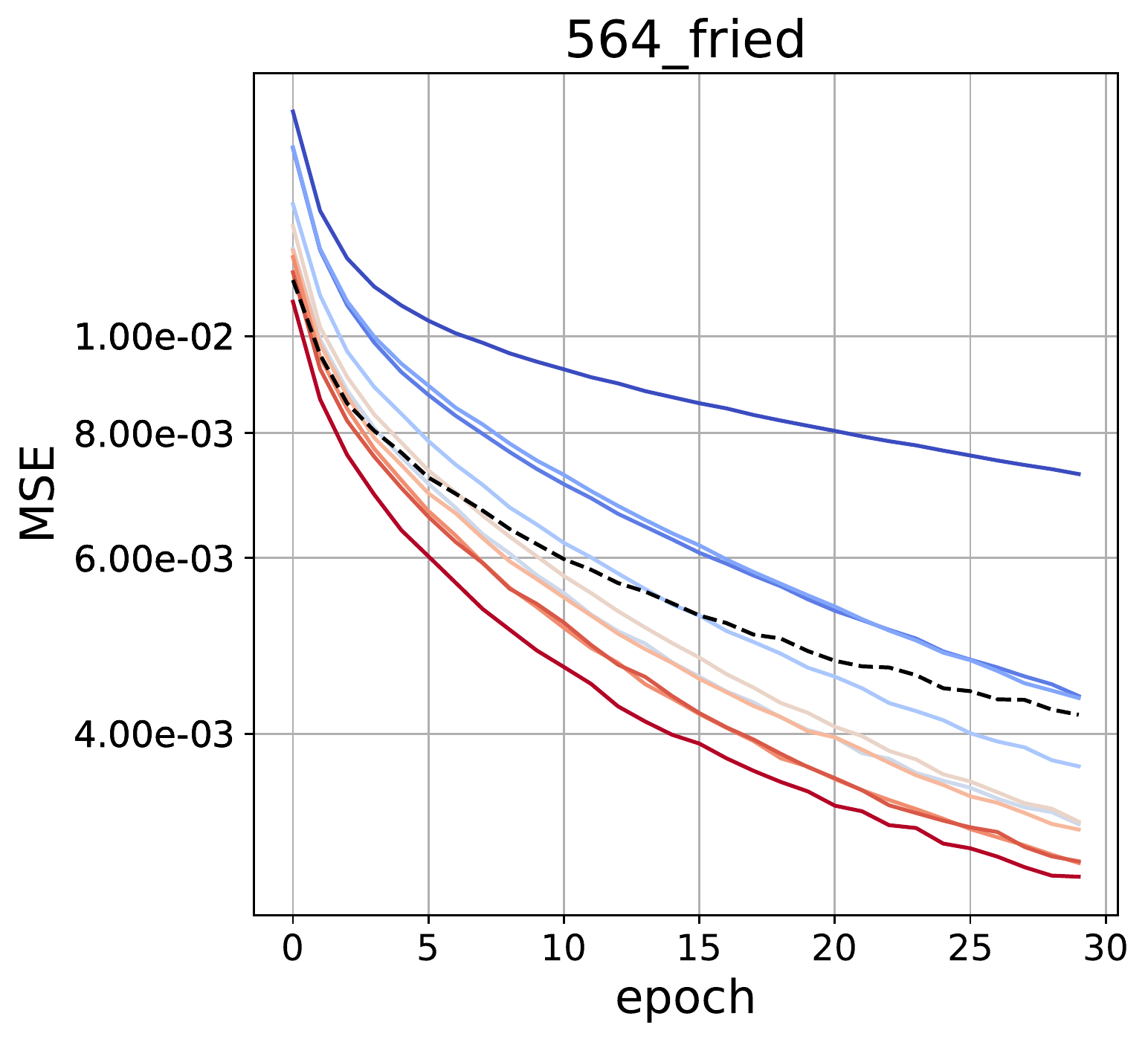}
\includegraphics[height=4.8cm]{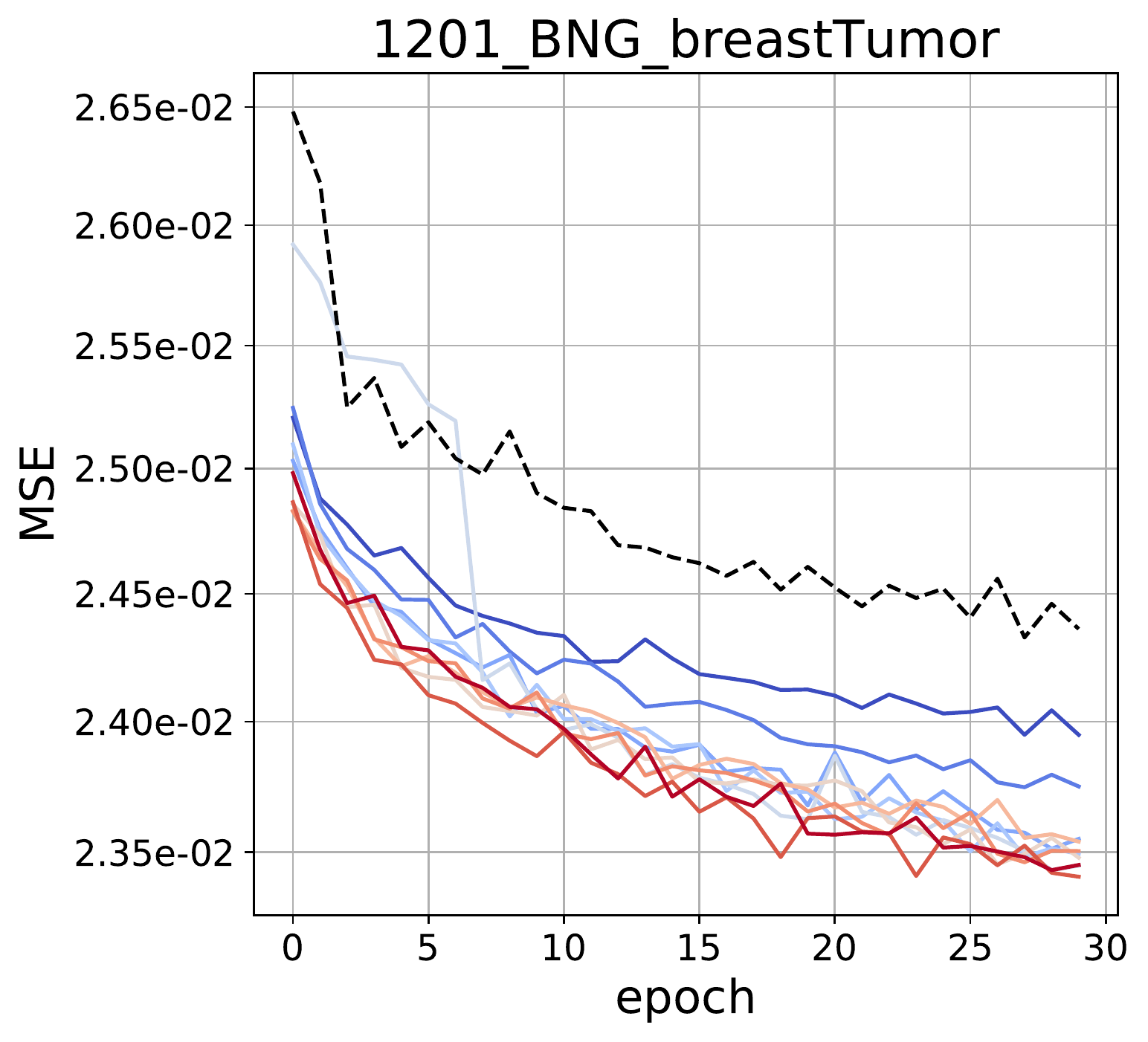}
\caption{Learning curves for selected regression problems, starting with a 4 layer feed-forward neural network (dark blue) and applying the LSMF-algorithm for 10 iterations (last iteration: dark red). Reported is the average MSE of a population of 100 dCGPANN with different sets of starting weights in log-scale. The dashed black line corresponds to the average MSE of a population of 100 randomly generated dCGPANNs, each running with one of the 100 starting weights.}
\label{fig:evoruns}
\end{figure*}

Our experiments show that LSMF is able to lower the average test error after each evolutionary iterations for all selected problems. In Figure~\ref{fig:evoruns} we show the test error for each evolutionary run averaged over a population of 100 different random weight initializations. Each solid line corresponds to an unperturbed training (no mutations) of the best evolved topology together at the end of the previous iteration (and the template network for the first iteration).

To analyze if LSMF has any selective pressure to drive optimization towards better learning or simply amounts to some form of random search, we also visualize the average test error of $100$ random dCGPANNs, removing 5\% outlier. A random dCGPANN is generated by drawing random numbers uniformly for all genes within their corresponding bounds and constraints of $\mathbf{x_I}$ and by initializing $\mathbf{x_R}$ in the same way as non-random dCGPANNs (zero mean, normally distributed weights).

It turns out that after at most six evolutionary iterations, LSMF has evolved topologies which perform better than random dCGPANNs on average. The difference between the test error of the randoms networks in comparison with the test error of the best evolved topology is significant with $p < 0.05$ according to separate Wilcoxon Rank sum tests for each regression problem. Remarkably, the random dCGPANNs seem to perform (on average) still better than the initial feed-forward neural networks. The (average) performance of the random dCGPANNs is shown as dashed black line in Figure~\ref{fig:evoruns}. Numerical values for training and test error of dCGPANNs and the test error of the random dCGPANNs are shown in Table~\ref{tab:funny_numbers}.

\begin{figure*}[tbp]
\centering
\includegraphics[width=3.5cm]{img/network0.pdf}
\includegraphics[width=3.5cm]{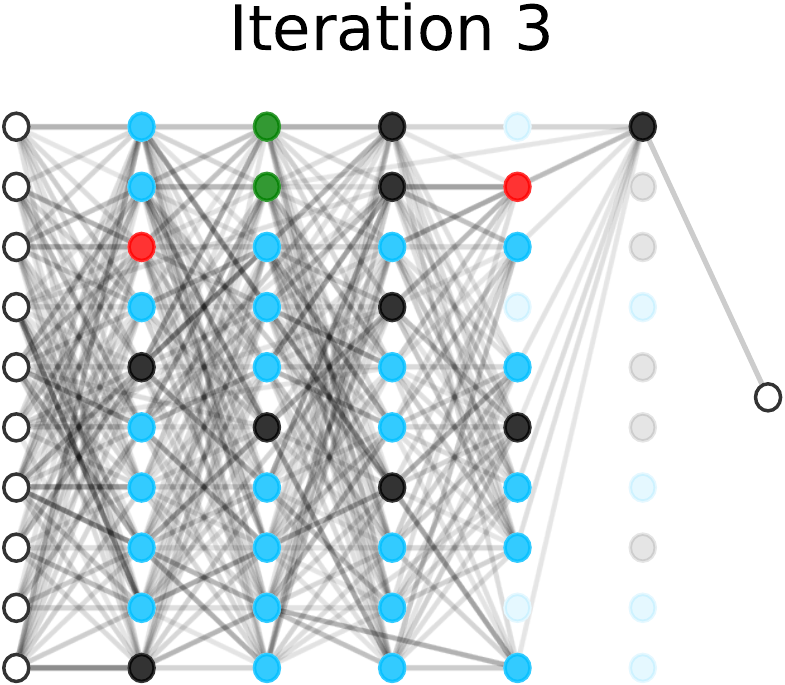}
\includegraphics[width=3.5cm]{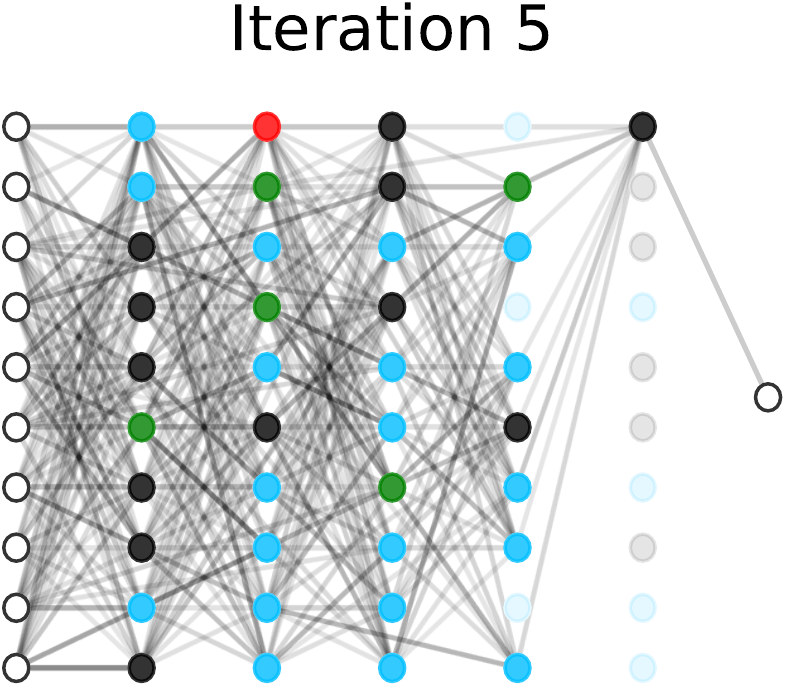}
\includegraphics[width=3.5cm]{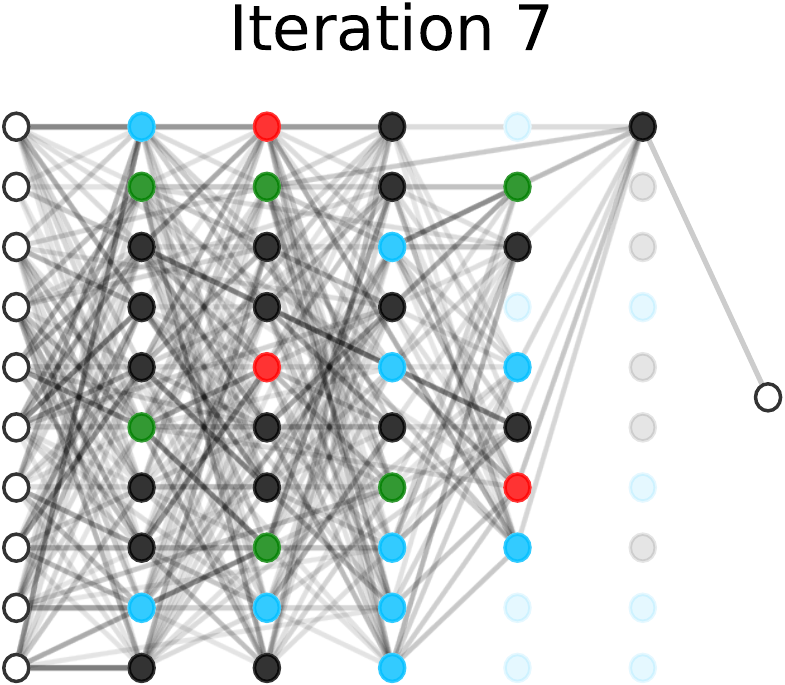}
\includegraphics[width=3.5cm]{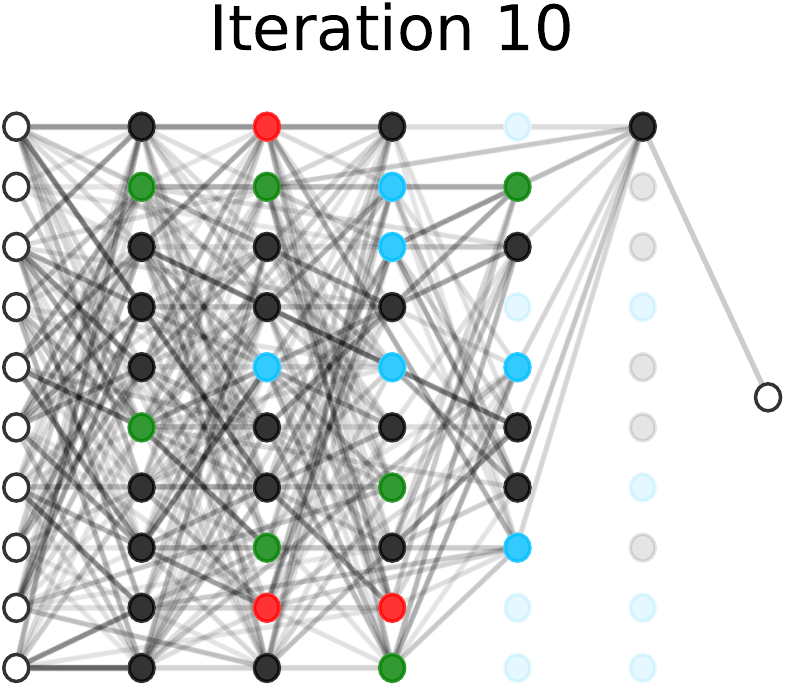}\\
\includegraphics[width=10.0cm]{img/legend.pdf}
\caption{Evolution of dCGPANN topology by running the LSMF-algorithm on the problem 344\_mv. Leftmost: initial topology (feed-forward neural network). From left to right: best performing topologies (lowest MSE at the end of an iteration) at several points during the 10 iterations of LSMF. The thickness of a connection is proportional to its weight.}
\label{fig:network_evo}
\end{figure*}

Figure~\ref{fig:network_evo} shows an example of an evolving topology by depicting the network structure of the dCGPANN with lowest training error at the end of each corresponding iteration. For this particular example, the impact of the activation function mutations is clearly visible: the prevalence of \texttt{tanh} nodes in the hidden layers drops from 100\% down to approx. 13\%, which is much lower than one would expect by a purely a random assignment of activation functions. We further observe that sigmoidal activations appear dominantly, next to \texttt{ReLU} and occasional \texttt{ELU} nodes. Moreover, the \texttt{tanh} activation functions of the first layer are completely substituted after the last evolutionary iteration.

Analyzing the structure of the evolved network population, we observe across all problems that certain connections are dropped while others are enforced by rewiring links on top of each other. While the dCGPANN encoding enables such $k$-fold links, they are redundant for computation and may be substituted by a single link containing the sum of the $k$ connection weights.

\begin{figure}[btp]
\centering
\includegraphics[width=0.23\textwidth, trim={0.6cm 0 0 0}, clip]{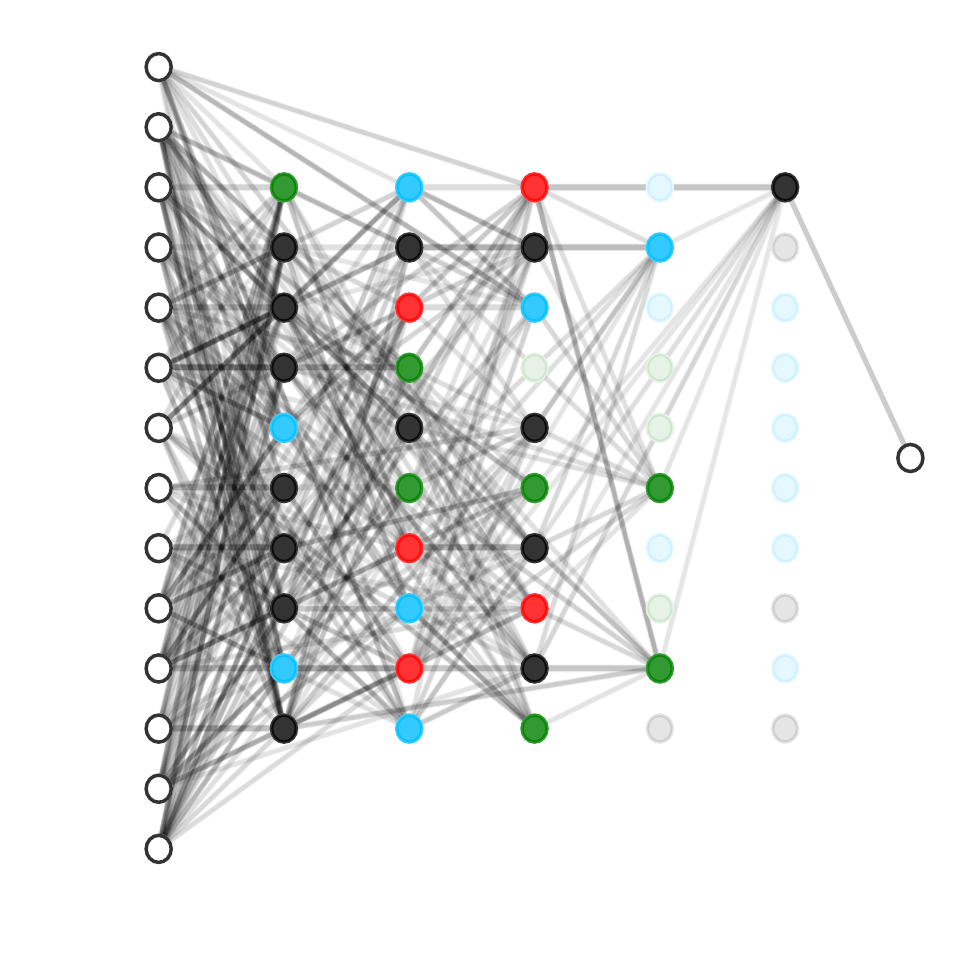}
\includegraphics[width=0.23\textwidth, trim={0.6cm 0 0 0}, clip]{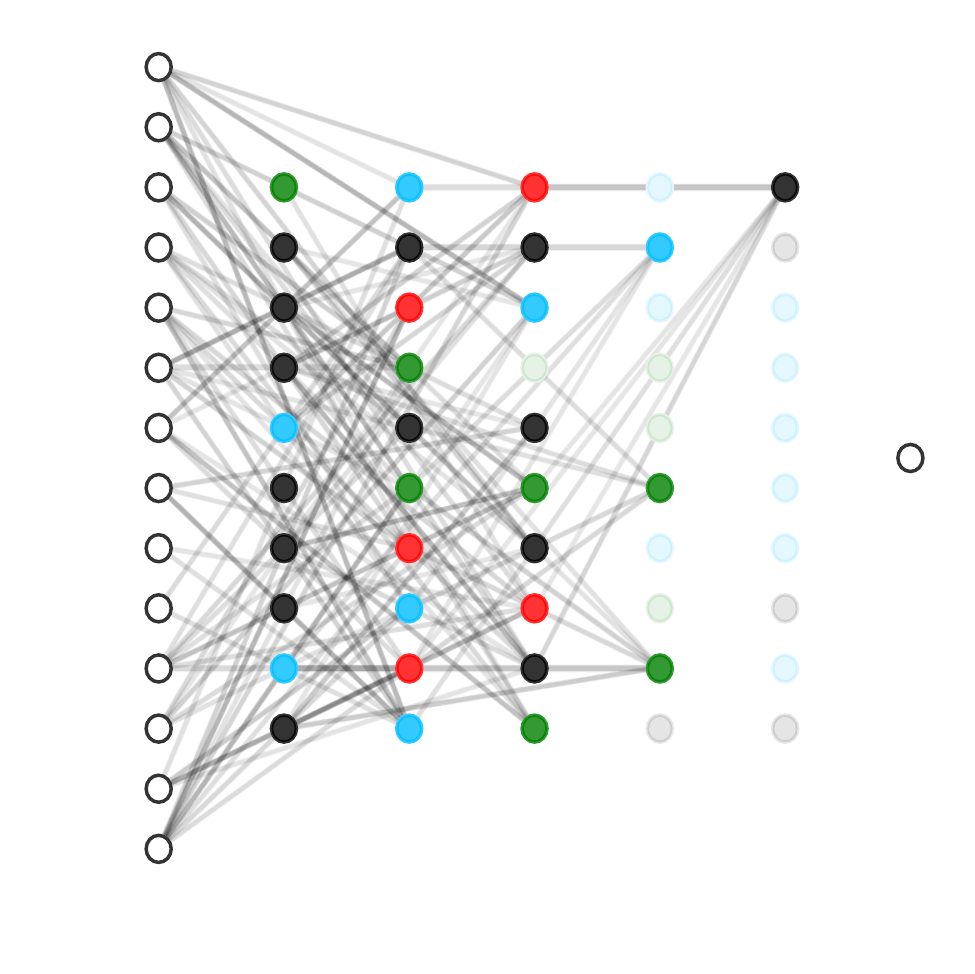}
\includegraphics[width=0.45\textwidth]{img/legend.pdf}
\caption{Final evolved topology after 10 iterations of LSMF on the problem 503\_wind. Left: all active connections. Right: skip connections only (active connections who do not connect two nodes in neighboring columns).}
\label{fig:windnet10}
\end{figure}

We further observe that, on average, half of the nodes in the last layer become inactive, which further reduces the size of the network model. Only for the 503\_wind problem, nodes in lower layers have been observed to become inactive as well (see Figure~\ref{fig:windnet10}). Inactive connections and duplicate connections allow to lower the space requirements for the models by about 40\%. Specific values for possible model compression are listed in Table~\ref{tab:compression_skip}.

The reason for the appearance of inactive nodes are rewired connections that go either into $k$-fold links or to an entirely different hidden layer. As the levels-back parameter in our experiment is $3$, any of the $10$ incoming connections of the output node can be rewired to \emph{skip} the last (or even the second-last) layer, necessarily resulting in inactive nodes within the last layer. These skip connections can appear also at other places in the network, bypassing other hidden layers or feeding directly into the inputs (compare right side of Figure~\ref{fig:windnet10}). We report the total number of evolved skip-connections in Table~\ref{tab:compression_skip}.

\section{Discussions and Conclusion}
\label{sec:discussion}

Our experiments show that it is possible to find a dCGPANN starting from a feed forward neural network that increases the speed of learning while at the same time reducing the complexity of the model for several regression problems. These two effects might not be unrelated, as smaller models are (generally) faster to train. However, while random dCGPANNs are on average even smaller than the evolved dCGPANNs, their performance falls behind after a couple of evolutionary iterations. This implies that LSMF-like algorithms are able to effectively explore the search space of dCGPANN topologies. Thus, there are reasons to assume that the performance of neural networks might be generally enhanced by the deployment of dCGPANNs.

For example, when comparing the evolved structures, it becomes clear that certain activation functions are selected more than others. This is reflected in recent neural network research, which acknowledges their importance~\cite{basirat2018quest, ramachandran2018searching, Nair2010Rectified}. While so far only a few vague rules and human intuition could assist in setting these functions in favor of particular problems, LSMF allows to automatically evolve beneficial activation functions on the level of individual neurons.

Another reason for the superior performance of dCGPANNs is the emergence of skip connections, which are widely acknowledged as one of the major mechanism to overcome the vanishing gradient problem~\cite{hochreiter1998vanishing}, a major roadblock in the development of Deep Learning. Many of the modern deep architectures like ResNets~\cite{He2016Deep}, HighwayNets~\cite{Srivastava2015Highway} and DenseNets~\cite{Huang2017DenselyCC} need such skip connections to enable effective training.

A further well-known method that has been used to enhance neural network training is dropout~{\cite{Srivastava2014Dropout}. This procedure is simulated by dCGPANN by use of inactive nodes, which are frequently appearing and disappearing due to the random rewirings of Mutate steps. Dropout is mainly used to prevent overfitting, a problem that we have not encountered in our experiments so far. However, if overfitting was an issue, a three-way split of the data would provide an easy solution as it would enable the Select step to be independent from the training and test data.

While 10 iterations of LSMF have been proven effective for most problems, there is no guarantee that the algorithm will \emph{always} reduce the error after every iteration (In fact, the problem 344\_mv in Figure~\ref{fig:evoruns} provides a counterexample). This convergence behavior could arise from the rigid use of hyperparameters that we deployed to show the broad applicability of LSMF. A problem-dependent adaptation of these hyperparameters between different iterations of LSMF would most likely mitigate this issue.

Due to an efficient implementation of automated differentiation, the resources which are needed to enable backpropagation in dCGPANNs are negligible. In fact, the software we deploy evaluates (on the CPU) networks with their gradient nearly 2 times faster than comparable neural network frameworks like tensorflow or pytorch on the problems that we presented here. The emergence of skip connections and dropout-like behaviour gives us reason to believe that evolution of dCGPANNs might be especially beneficial for larger and deep architectures. In particular, any pre-defined deep layered network can be transformed into a dCGPANN to serve as a starting point for structural optimization while training takes place. 
Also smaller dCGPANN subnetworks (i.e. last layers) can be deployed for evolution or used as buildings blocks inside automated machine learning frameworks (e.g.~\cite{guyon2016brief}). However, because in its current release no SIMD vectorization nor GPU support is available, analyzing larger network architectures (as they are frequently deployed for example in image classification tasks), becomes very time consuming. Thus, the scalability of our approach has to remain open until GPU-support or equivalent forms of parallelization become available for proceeding studies.


Next to matters of scalability, further research can be directed to more sophisticated genetic algorithms working on dCGPANNs. While our mutations targeted connections and activation functions, it would be of interest to study their effects separately and further develop mutation operators (potentially by incorporating gradients) that further improve the algorithm. While our choices of hyperparameters achieved already promising results, a better adaptation of the cooldown period, learning rate, mutation rates and weight initialization could potentially lead to even larger improvements. We deliberately avoided delving too deep into the impact of these hyperparameters to demonstrate the broad applicability of evolutionary optimization in combination with stochastic gradient descent for weight adaptation. Since both of these aspect can be handled efficiently if neural networks are represented as dCGPANNs, we believe that this encoding can open the door to transfer plenty of helpful and established evolutionary mechanism into network training, bringing us a step closer to achieving neural plasticity for our artificial neural networks.

\bibliographystyle{ACM-Reference-Format}
\bibliography{dcgpann} 

\end{document}